\documentclass[10pt,twocolumn,letterpaper]{article}

\usepackage{cvpr}              

\usepackage[dvipsnames]{xcolor}

\definecolor{cvprblue}{rgb}{0.21,0.49,0.74}
\usepackage{natbib}
\usepackage{amsmath}
\usepackage{amsfonts}
\usepackage{setspace}
\usepackage{float}
\usepackage{relsize}
\usepackage{graphicx}
\usepackage{color}
\usepackage{dsfont}
\usepackage{subcaption}
\usepackage{svg}
\usepackage[pagebackref,breaklinks,colorlinks,citecolor=cvprblue]{hyperref}

\def\paperID{21} 
\def\confName{CVPR}
\def\confYear{2024}

\title{The Penalized Inverse Probability Measure for Conformal Classification}

\author{Paul Melki\\
IMS, CNRS, University of Bordeaux\\
EXXACT Robotics \\
{\tt\small paul.melki@u-bordeaux.fr}
\and
Lionel Bombrun\\
IMS, CNRS, University of Bordeaux\\
Bordeaux Sciences Agro\\
{\tt\small lionel.bombrun@ims-bordeaux.fr}\\
\and
Boubacar Diallo, Jérôme Dias\\
EXXACT Robotics \\
{\tt\small boubacar.diallo@exxact-robotics.com}\\
{\tt\small jerome.dias@exxact-robotics.com}
\and
Jean-Pierre Da Costa\\
IMS, CNRS, University of Bordeaux\\
Bordeaux Sciences Agro\\
{\tt\small jean-pierre.dacosta@ims-bordeaux.fr}\\
}

\begin{document}

\raggedbottom

\maketitle
\begin{abstract}
    The deployment of safe and trustworthy machine learning systems, and particularly complex black box neural networks, in real-world applications requires reliable and certified guarantees on their performance. The conformal prediction framework offers such formal guarantees by transforming any point into a \textit{set predictor} with valid, finite-set, guarantees on the coverage of the true at a chosen level of confidence. Central to this methodology is the notion of the nonconformity score function that assigns to each example a measure of ``strangeness" in comparison with the previously seen observations. While the coverage guarantees are maintained regardless of the nonconformity measure, the point predictor and the dataset, previous research has shown that the performance of a conformal model, as measured by its efficiency (the average size of the predicted sets) and its informativeness (the proportion of prediction sets that are singletons), is influenced by the choice of the nonconformity score function. The current work introduces the \emph{Penalized Inverse Probability (PIP)} nonconformity score, and its regularized version \emph{RePIP}, that allow the joint optimization of both efficiency and informativeness. Through toy examples and empirical results on the task of crop and weed image classification in agricultural robotics, the current work shows how PIP-based conformal classifiers exhibit precisely the desired behavior in comparison with other nonconformity measures and strike a good balance between informativeness and efficiency.
    
\end{abstract}
\vspace{-0.4cm}

\section{Introduction}
The development and deployment of machine learning-based autonomous systems has been a flourishing field of research in both academia and, relatively more recently, in the industry \citep{paleyesChallengesDeployingMachine2023, huangSurveySafetyTrustworthiness2020}. While machine learning models often exhibit high performance ``in the lab", they often face much more difficulty when deployed in the real world, for a number of reasons that are not yet fully clear \citep{damourUnderspecificationPresentsChallenges2022}. Indeed, when faced with a new observation, the model will produce a new prediction whose quality is often related to the similarity of this new observation to what the model has previously seen. When the new observation is quite anomalous with respect to the previously seen data or even slightly perturbed, most models will produce wrong predictions \citep{szegedyIntriguingPropertiesNeural2014a}, with often dire and intolerable consequences in safety critical applications such as autonomous driving \citep{spanfelner2012challenges, raoDeepLearningSelfdriving2018, alamAdversarialExamplesSelfDriving2022} and medical diagnosis \citep{petersenResponsible, pricePotentialLiabilityPhysicians2019, malihaArtificialIntelligenceLiability2021}, to name a few. 

The safe deployment of machine learning systems in the real world is therefore incumbent upon the integration of at least two main important features into them \citep{hendrycksUnsolvedProblemsML2022, huangSurveySafetyTrustworthiness2020}: (1) the ability to provide valid and trustworthy guarantees on the quality of predictions in ``normal" conditions, and (2) the ability to reliably detect and signal anomalies when faced with them.

Conformal prediction is a method that provides formal statistical guarantees on the predictive quality of any black box model \citep{vovkAlgorithmicLearningRandom2005, angelopoulosConformalPredictionGentle2023a}. It has recently gained in popularity due to the minimal assumptions required for its deployment. Without imposing explicit conditions on the data distribution, any base point predictor can be transformed using the conformal approach into a set predictor with formal guarantees on the coverage of the true value at confidence level $1-\alpha$, where $\alpha$ is a chosen level of tolerance to error. Formally, in a supervised learning context, whereby for each object $\mathbf{x} \in \mathcal{X}$ is assigned a label $y \in \mathcal{Y}$, a conformal model produces prediction sets $\mathcal{C}_{1-\alpha} \subset \mathcal{Y}$ that satisfy the \textit{marginal coverage} guarantee \citep{angelopoulosConformalPredictionGentle2023a, romanoClassificationValidAdaptive2020}
\begin{equation} \label{eq:marg_coverage}
    \mathbb{P} \big(y \in \mathcal{C}_{1 - \alpha}(\mathbf{x}) \big) \ge 1 - \alpha
\end{equation}
whenever the test data follow the same distribution as the data on which the model was calibrated. Under this condition, the coverage guarantee is satisfied marginally over all possible calibration sets. Additionally, the study of the structure and the size of the predicted sets allows us to quantify the uncertainty of the base model, and to detect examples on which the model is highly uncertain \citep{laxhammarChapterAnomalyDetection2014}. As such, the conformal approach can be used to satisfy the two conditions for safe deployment of machine learning systems as it has been shown in a number of applications \citep{balasubramanianConformalPredictionReliable2014a} ranging from railway signaling \citep{andeolConfidentObjectDetection2023}, medical imaging \citep{luFairConformalPredictors2022}, to nuclear fusion \citep{vegaAccurateReliableImage2010}.

Three main components are needed to conduct inductive conformal prediction \citep{papadopoulosInductiveConfidenceMachines2002}: a base predictor $\mathcal{B}$ (which can be any machine learning point predictor), a dataset on which to calibrate the model so that it becomes a conformal predictor, and a nonconformity score function $\Delta$ that assigns a ``strangeness value" to each example in the calibration set. This value measures how \textit{conforming} each individual is to what the model has previously seen. While the marginal coverage guarantee is satisfied by construction, the quality of the predicted sets is influenced by these three components. For example, a neural network $\mathcal{B}$ with low accuracy can still be calibrated to achieve $1 - \alpha = 0.9$ coverage, but will tend to predict much larger sets, since it is uncertain about the true class and thus needs to predict many to guarantee the inclusion of the true one.

The object of interest in this work is the nonconformity score function $\Delta$. In particular, we are interested in studying the influence of different nonconformity functions on two of the most commonly used metrics for the evaluation of conformal classifiers \citep{johanssonModelagnosticNonconformityFunctions2017}: \textit{efficiency}, the average size of the predicted sets, and \textit{informativeness}, the proportion of predicted singleton sets. These two metrics measure, in some sense, the ``usability" of the conformal approach when needed to take decisions under normal condition, and may be useful to signal high uncertainty conditions. The context of the study is automated precision weeding in agriculture \citep{colemanWeedDetectionWeed2022}, whereby a robotic system is embedded on a tractor to detect and spray herbicides on undesirable weeds in real-time, under real-world conditions. The precision agriculture sector is an interesting test-bed for safe AI methodologies since they are indeed needed in agriculture, but do not directly threaten human lives in case of failure.

\vspace{0.2cm}
\noindent \textbf{Related work} \quad A good body of research is dedicated to the development of useful and efficient nonconformity score functions \citep{bostromEvaluationVarianceBasedNonconformity2016, linussonEfficiencyComparisonUnstable2014, gaurahaSynergyConformalPrediction2021a}. For classification, the first comprehensive work is that of Johansson \textit{et al.} \citep{johanssonModelagnosticNonconformityFunctions2017} in which the authors study the impact of different model-agnostic nonconformity functions -- in particular, the Hinge Loss and the Margin Score -- on neural network classifiers. The authors find that neither of these score functions allows the joint maximization of informativeness and efficiency. Their empirical results show that the Hinge Loss minimizes the size of prediction sets, while the Margin Score maximizes the number of singletons. These results are further confirmed by Aleksandrova and Chertov~\cite{aleksandrovaHowNonconformityFunctions2021, aleksandrovaImpactOfModelAgnostic} on most of the datasets they tested, in their work aiming at reconciling the two scores by computing, for a new observation, two conformal sets using both the Hinge and Margin scores, then choosing the Margin-based set as the final prediction if it is a singleton, or the Hinge set otherwise. Unfortunately, this approach may be quite inefficient as it requires repeating the calibration step for each nonconformity function. Fisch \textit{et al.} \citep{fisch2020efficient} propose an efficient conformal classification approach based on an expansion of the notion of validity to include the concept of \textit{admissible} labels, which are semantically plausible class labels for a given example. Such an expansion may lead to highly inefficient prediction sets in learning tasks with a large number of classes. As such, the authors develop an efficient cascaded inference algorithm that reduces the size of the prediction set by progressively filtering the number of candidates via a sequence of increasingly complex classifiers. Other works have explored ways to combine multiple conformal models in such a way as to preserve the validity guarantee while producing sets that are as efficient as possible \citep{toccaceliCombinationConformalPredictors2017, toccaceliCombinationInductiveMondrian2019, gaurahaSynergyConformalPrediction2021a}.

\vspace{0.2cm}
\noindent \textbf{Contributions} In direct continuation of these previous works, and for the expansion of the still meager body of work on conformal prediction in precision agriculture \citep{faragInductiveConformalPrediction2023, melkiGroupConditionalConformalPrediction2023a, chiranjeeviDeepLearningPowered2023}, our work proposes the following contributions:
\begin{enumerate}
    \item The proposal of a new model-agnostic nonconformity function that strikes a good balance between optimizing both efficiency and informativeness: the \textit{Penalized Inverse Probability} (PIP);
    \item The proposal of a simple regularized version of PIP, RePIP, inspired by \citep{angelopoulosUncertaintySetsImage2021} for improved efficiency in use cases with a large number of classes;
    \item The comparison of PIP with other nonconformity measures from the literature on toy examples, showing the balanced and adaptive behavior of this measure under different settings;
    \item The comparison of PIP and RePIP with other nonconformity measures from the literature based on efficiency and informativeness through rigorous empirical experiments on an image dataset for crop and weed classification taken under real-world conditions with the aim of providing valid guarantees on the performance of a precision weeding system.
\end{enumerate}

\vspace{-0.1cm}
\section{Definitions \& Mathematical Setup}
Let $\mathbf{x} \in \mathcal{X}$ be a vector of features, which we will call an \textit{object} \cite{vovkAlgorithmicLearningRandom2005}. To each object is associated a class label $y \in \mathcal{Y} := \{1,..., K\}$ to form what we call an \textit{example} $\mathbf{z} = (\mathbf{x}, y) \in \mathcal{X} \times \mathcal{Y}$. A black-box classifier $\mathcal{B}$ is trained on a set of $n_{\text{train}}$ examples to output for an object a class prediction $\hat{\mathcal{B}}(\mathbf{x}) = \hat{y} \in \{1,..., K\}$ and an associated estimated probability $\hat{p}^{\hat{y}} \in [0, 1]$, such that $\sum_{k=1}^K \hat{p}^k = 1$.

The inductive conformal approach consists of a calibration step in which the trained classifier is calibrated on a set of $n_{\text{cal}}$ calibration examples $\{\mathbf{z}_i = (\mathbf{x}_i, y_i), i = 1, ..., n_{\text{cal}} \}$ using a real-valued nonconformity score function $\Delta(\mathbf{z}): \mathcal{X} \times \mathcal{Y} \rightarrow \mathbb{R}$. The output of the calibration step is usually a quantile value $q_{\text{cal}} \in \mathbb{R}$ computed on the distribution of nonconformity scores over the calibration set. 

This quantile is then used to produce prediction sets $\mathcal{C}_{1-\alpha} (\mathbf{x}) \subset \mathcal{Y}$ on the remaining $n_{\text{test}}$ test examples. For each class, its score $\Delta$ is computed based on the probability estimated by $\mathcal{B}$, then compared to $q_{\text{cal}}$ in a hypothesis test of whether the class is considered ``conforming" enough or not. The produced prediction sets are \textit{valid} in the sense that they satisfy the marginal coverage guarantee defined in Equation (\ref{eq:marg_coverage}). This property is verified empirically by computing the \textbf{empirical marginal coverage}, which is simply the proportion of prediction sets that cover the true label: 
\begin{equation}
    \frac{1}{n_{\text{test}}} \sum_{i=1}^{n_{\text{test}}}  \mathds{1}_{\{ y_i \in \mathcal{C}_{1 - \alpha}(\mathbf{x}_i) \}}
\end{equation}

The quality of the prediction sets can then be evaluated using these two metrics:
\begin{itemize}
    \item \textbf{Efficiency}, defined as the average size of the predicted sets: 
    \begin{equation} \label{eq:efficiency}
        \frac{1}{n_{\text{test}}} \sum_{i=1}^{n_{\text{test}}} | \mathcal{C}_{1 - \alpha}(\mathbf{x}_i) |
    \end{equation}
    where $|~.~|$ is the set cardinality, the number of classes in the predicted set.

    \item \textbf{Informativeness}, defined as the percentage of predicted sets of size 1 (often called \textit{oneC} in the literature \citep{johanssonModelagnosticNonconformityFunctions2017, aleksandrovaHowNonconformityFunctions2021}):
    \begin{equation} \label{eq:informativeness}
        \frac{1}{n_{\text{test}}} \sum_{i=1}^{n_{\text{test}}}  \mathds{1}_{\{| \mathcal{C}_{1 - \alpha}(\mathbf{x}_i)|  = 1\}}
    \end{equation}
\end{itemize}
Clearly, conformal predictors that have both high efficiency and high informativeness are the preferred models in practice, at a fixed coverage level of $1- \alpha$. Smaller set sizes are easier to manipulate and be used to construct decision rules. Singleton predictions are the most informative predictions since they do not manifest any ``uncertainty" about the predicted class. A most informative, and efficient, conformal model would be one that predicts only singletons while guaranteeing marginal coverage. Unfortunately, such an optimal conformal model is impossible to attain in practice \citep{romanoClassificationValidAdaptive2020}.

\section{Nonconformity Score Functions}
\subsection{Review of Some Nonconformity Scores}
The nonconformity measure quantifies the ``strangeness" of a given object by comparing it to the objects previously encountered by the model during training and calibration~\citep{shaferTutorialConformalPrediction2008}. For the same base predictor $\mathcal{B}$, different nonconformity functions lead to different conformal predictors. Here, we review commonly used nonconformity score functions for classification from the literature \citep{johanssonModelagnosticNonconformityFunctions2017, romanoClassificationValidAdaptive2020}. Since the estimated probabilities $\hat{p}^k$ are fixed for a given object $\mathbf{x}$, the nonconformity score function $\Delta(\mathbf{z})$ will simply be denoted $\Delta(y)$ in the following for ease of understanding. Note also that during the calibration step of the conformal procedure, $y$ is the true class of object $\mathbf{x}$, while during the prediction phase, $y$ is the tested class to be included or not in the prediction set.

\vspace{0.2cm}
\noindent  \textbf{Hinge Loss (IP)} \citep{johanssonModelagnosticNonconformityFunctions2017} \quad Also known as \textit{Inverse Probability}, this score function measures how far the estimated probability of $y$ (where $y$ is the true class label) is from the perfect score of 1: 
\begin{equation} \label{eq:hinge}
        \Delta^{\text{IP}}(y) = 1 - \hat{p}^y
\end{equation}
Indeed, a perfect classifier should always assign a probability of 1 to the true class label, which would have a Hinge score of 0. For smaller probability estimates of $y$, a higher Hinge score is assigned since the model is deemed more uncertain about $y$. The Hinge Loss can thus be considered a very ``natural" measure of nonconformity. Unfortunately, it suffers from a major shortcoming: it does not take the probability estimates of the other classes into consideration. 

\vspace{0.2cm}
\noindent \textbf{Margin Score (MS)} \cite{johanssonModelagnosticNonconformityFunctions2017} \quad Assuming an implicit hypothesis that good predictive models should assign the highest probability estimate to the true class, the MS measures the difference between the estimated probability of $y$ and the highest estimated probability among the other classes:
\begin{equation} \label{eq:margin}
    \Delta^{\text{MS}}(y) = \max_{k \ne y} \hat{p}^k - \hat{p}^y = \Delta^{\text{IP}}(y) +  \underbrace{\max_{k \ne y} \hat{p}^k - 1}_{\text{penalization}}
\end{equation}
A large positive value of this score indicates that the estimated probability assigned to $y$ is distant from the class of highest confidence. It means that class $y$ is considered highly strange in comparison to the class the model considers as the true one. Notice that $y$ is \textit{always} penalized, even when it is the most probable class, which is not ideal. Another shortcoming of the MS is that it only takes the maximum probability into consideration, why not take the probabilities of the other classes directly into consideration?  It is important to note that in cases of anomalies, OOD observations or adversarial attacks, neural networks would tend to assign the highest confidence to classes that are completely wrong \citep{szegedyIntriguingPropertiesNeural2014a}, thus putting the reliability of the Margin Score into question. 

\vspace{0.2cm}
\noindent \textbf{Regularized Adaptive Prediction Sets (RAPS)} \quad This nonconformity function was first introduced in \citep{romanoClassificationValidAdaptive2020} as part of the APS approach, with the aim of producing prediction sets whose size adapts to, and reflects, the difficulty of each object. It is the first score function that fully integrates a range of estimated probabilities other than that of $y$. In particular, the APS score incorporates all the estimated probabilities that are larger than that of the class of interest. Observing that the APS score tends to predict relatively large set sizes in learning problems with a large number of classes, Angelopoulos \textit{et al.} \citep{angelopoulosUncertaintySetsImage2021} introduced a regularized version of this score, named RAPS.

Let the operator $R(k)$ be the rank of class $k$ after the estimated probabilities $p^1, ..., p^K$ have been sorted in decreasing order, and $\hat{p}^{[r]}$ be the probability estimate of the class having rank $r$, such that $\hat{p}^k = \hat{p}^{[R(k)]}$, we can define the RAPS score function as:
\begin{equation} \label{eq:raps}
    \resizebox{.9\hsize}{!}{$    \Delta^{\text{RAPS}}(y) = \underbrace{\sum_{r = 1}^{R(y)-1} \hat{p}^{[r]} + u \cdot \hat{p}^{[R(y)]}}_{\text{APS}}  + \underbrace{\lambda \cdot \big(R(y) - k_{reg} \big)^+}_{\text{regularization}}$}
\end{equation}
where $u$ is a uniform random variable in $(0, 1)$ for tie-breaking, $\lambda$ is the penalization amount and $k_{reg}$ is the rank at which to start penalizing. $\lambda$ and $k_{reg}$ can be fixed by the user or optimized on a held-out dataset. The penalization is proportional to the how further away is $y$ in the ranking of estimated probabilities from $k_{reg}$. When $y$ has a very low probability, meaning that it has a very high $R(y)$, its score will be very strongly penalized, leading to the exclusion of $y$ from the prediction set. This will lead, on average, to smaller set sizes, as it excludes from the prediction sets those classes that would have been included by the original APS score (obtained for $\lambda = 0$). While the APS and RAPS have been developed with adaptivity and efficiency in mind, their authors' do not seem to take the informativeness criterion into consideration.

\subsection{Penalized Inverse Probability}
In this article, we introduce the \textit{Penalized Inverse Probability} (PIP), a new nonconformity score function that integrates components from the three previously presented measures with the aim of optimizing both efficiency and informativeness. Following the same notation presented previously, PIP can be defined as:
\begin{equation} \label{eq:pip}
    \Delta^{\text{PIP}}(y) = \underbrace{1 - \hat{p}^y}_{\Delta^{\text{IP}}(y)} + \underbrace{ \sum_{r = 1}^{R(y)-1} \frac{\hat{p}^{[r]}}{r} ~\mathds{1}_{\{R(y) > 1 \}}}_{\text{penalization}}
\end{equation}
For $R(y) = 1$, when $y$ is the class with the highest estimated probability, the PIP score is simply the Hinge (IP) Loss. In all other cases, the sum of the estimated probabilities of all the classes with higher probability than $y$ weighted by the inverse of their rank is added as a penalization term. As such, a decreasing weight is associated to each class that is closer to $y$. This penalization term resembles the APS score, and alleviates the shortcoming inherent by IP of not taking the estimated probabilities of other classes into consideration. Furthermore, for $R(y) = 2$, it should be clear that $\Delta^{\text{PIP}}(y) = 1 + \Delta^{\text{MS}}(y)$. As such, the PIP score exhibits analogous behavior to different nonconformity functions depending on the estimated probabilities by the base model $\mathcal{B}$, leading to better adaptivity, as will be shown in the toy examples below. For more detailed developments on the relationship between the PIP and the other scores, we refer the interested reader to Section 1 in the Supplementary Material.

\begin{figure}[t!]
    \centering
    \includegraphics[width=\linewidth]{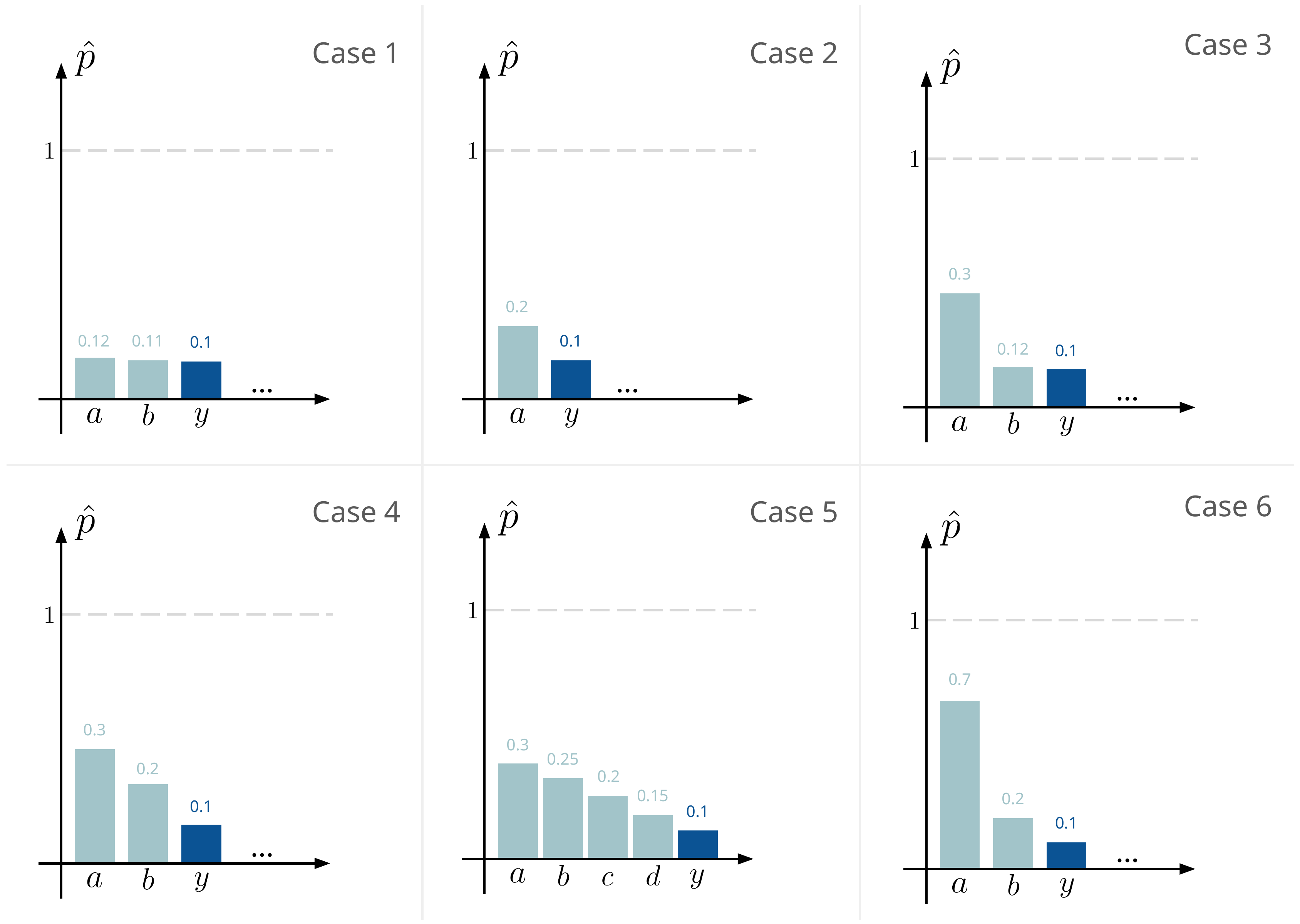}
    \caption{Six different potential configurations of model outputs sorted in decreasing order of $\hat{p}$. Only the classes until reaching the class of interest $y$ are shown. Computed nonconformity scores for each case can be seen in Table \ref{tab:scores_examples}.}
    \label{fig:softmax_examples}
\end{figure}

\begin{table}
    \centering
    \begin{tabular}{c|c|c|c|}
         & $\Delta^{\text{IP}}(y)$ & $\Delta^{\text{MS}}(y)$ & $\Delta^{\text{PIP}}(y)$ \\
         \hline 
         Case 1 & 0.90 & 0.02 & \textbf{1.08}\\
         Case 2 & 0.90 & 0.10 & \textbf{1.10}\\
         Case 3 & 0.90 & 0.20 & \textbf{1.26}\\
         Case 4 & 0.90 & 0.20 & \textbf{1.30} \\
         Case 5 & 0.90 & 0.20 & \textbf{1.43}\\
         Case 6 & 0.90 & 0.60 & \textbf{1.70}\\
                  
    \end{tabular}
    \caption{Computed scores of the example cases shown in \autoref{fig:softmax_examples}. The proposed $\Delta^{\text{PIP}}(y)$ manifests a more adaptive behavior for the varying configurations than the classical IP and MS functions.} 
    \label{tab:scores_examples}
\end{table}

\vspace{0.2cm}
\noindent \textbf{Toy examples} \quad  Consider the six different possible output configurations of a neural network classifier shown in \autoref{fig:softmax_examples}. The class of interest is $y$ and its estimated probability is fixed to $\hat{p}^y = 0.1$ in all the examples. Only the classes having higher estimated probabilities than $y$ are shown since they are the only ones that are used in the computations of the different scores. In~\autoref{tab:scores_examples} are shown the different scores assigned to class $y$ in each of the cases, sorted in increasing order. A greater score is a sign of greater ``nonconformity" -- that is, of higher uncertainty -- attributed to $y$. 

The first obvious observation is that IP assigns the same score to $y$ in all cases. As $\hat{p}^y = 0.1$ is the same in all cases and IP is, by definition, indifferent to the other classes, all the configurations are reduced to the same score. This rigidity is often undesirable in a nonconformity score function. 

The MS measure, on the other hand, manifests a more fluid behavior since it also considers the highest estimated probability. Case~1 is assigned the lowest MS score, since the estimated probabilities of $y$ and $a$ are quite similar. As such, MS considers that class $y$ is as likely a candidate as $a$ to be the first predicted class, and thus assigns it a low nonconformity score. Case~2 is considered a bit ``stranger" than Case~1 by the MS function because the difference between the maximal class $a$ and $y$ is a bit larger, which is a desirable behavior by this score function. In Case~6, although $y$ has the same rank $R(y) = 3$ as in Case~1, the MS value is maximal since the \textit{margin} between the $\hat{p}^a$ and $\hat{p}^y$ is large. Cases~3 to 5 show the shortcoming of the MS measure. Since in all these cases the difference between $\hat{p}^a$ and $\hat{p}^y$ is the same, they will all be assigned the same score value, even though it is clear that class $y$ in Case~5 should be assigned a higher nonconformity value than in Case 3 or even in Case 4. 

The proposed PIP function exhibits the most versatile behavior since it takes into consideration all the classes having higher estimated probabilities than $y$. $\Delta^{\text{PIP}}(y)$ is different in all the distinct configurations, manifesting the specificity of each case. Indeed, it can easily be shown that $\Delta^{\text{PIP}}$ guarantees a different score for every class even in the case of highest uncertainty where all the classes have the same estimated probability $1/K$ (Section 1 in Supplementary Material). Case~1 has the lowest PIP score, since class $y$ is almost as likely as $a$ or $b$ to be predicted as the first class. As such $y$ is not deemed strange in such a condition. The behavior of PIP in such situations is similar to that of MS. Case~2 is considered slightly stranger because the difference between $\hat{p}^a$ and $\hat{p}^y$ is larger and cannot simply be attributed to some ``noise." While $y$ has the same estimated probability and rank $R(y) = 3$ in both Case~3 and Case~4, it receives a slightly lower score in Case~3 since the difference with the $b$ is very small: $y$ could very much have been the second class and thus need not be penalized heavily for falling in third place. Class $y$ in Case~5 is further penalized because more significant classes have higher estimated probabilities than $y$.

\vspace{0.2cm}
\noindent \textbf{Summary of PIP score properties} \quad The desirable behavior manifested by PIP can be summarized as:
\begin{itemize}
    \item In all situations, the Hinge Loss (IP) is a baseline value for the PIP function. Therefore, classes with low probability estimates will tend to be assigned higher nonconformity scores. This kind of behavior leads to a lower average size of predicted sets (higher efficiency) since it tends to exclude the classes with low probability estimates \citep{johanssonModelagnosticNonconformityFunctions2017}.
    \item PIP takes into consideration all the probability estimates of the other classes with higher probabilities when computing the score for a given class. This includes the maximum probability class. Therefore, when $\hat{p}^y$ has a low value compared to $\max_{k \ne y} \hat{p}^k$, class $y$ will be heavily penalized (just like with the MS measure). This behavior generally leads to more predicted singletons (higher informativeness) because in all cases where one class has a very high probability estimate, all the other classes will be heavily penalized and thus excluded from the predicted set \citep{johanssonModelagnosticNonconformityFunctions2017}.
    \item Additionally, PIP distinguishes the cases where the difference between $\hat{p}^y$ and the ``more probable" classes is significant or not, penalizing less when such differences are negligible and can be attributed to some noise. This leads to scores that are different almost everywhere, permitting better discrimination between the different model outputs. 
\end{itemize}

\subsection{Regularized PIP}
For learning tasks with a large number of classes, the user may require to preserve the desirable behavior of the PIP score function but with smaller set sizes on average. The same regularization term added to APS \cite{angelopoulosUncertaintySetsImage2021} can be added to obtain RePIP, a regularized version of the proposed nonconformity measure:
\begin{equation}
    \Delta^{\text{RePIP}}(y) = \Delta^{\text{PIP}}(y) + \underbrace{\gamma \cdot \big(R(y) - k_{reg} \big)^+}_{\text{regularization}}
\end{equation}
Here, $\gamma$ is the equivalent of the $\lambda$ parameter in the RAPS nonconformity score and $k_{reg}$ is, similarly to RAPS, the rank at which to begin penalizing more. 

\section{Experimental Results}
In this section, we study the performance of different conformal classifiers obtained using the nonconformity score functions presented previously on the task of classifying images taken under real world conditions into 13 different plant species. This learning task is part of a precision weeding robotic use case, where an autonomous robot should distinguish weeds from cultivated crops and spray them with herbicide in real-time. Guaranteeing the performance of the weed classifiers is of great importance since missed weeds can multiply quickly and threaten heavily the health of the cultivated crops and the quality of harvest.

\begin{figure*}    
    \centering
    \includegraphics[width=\linewidth]{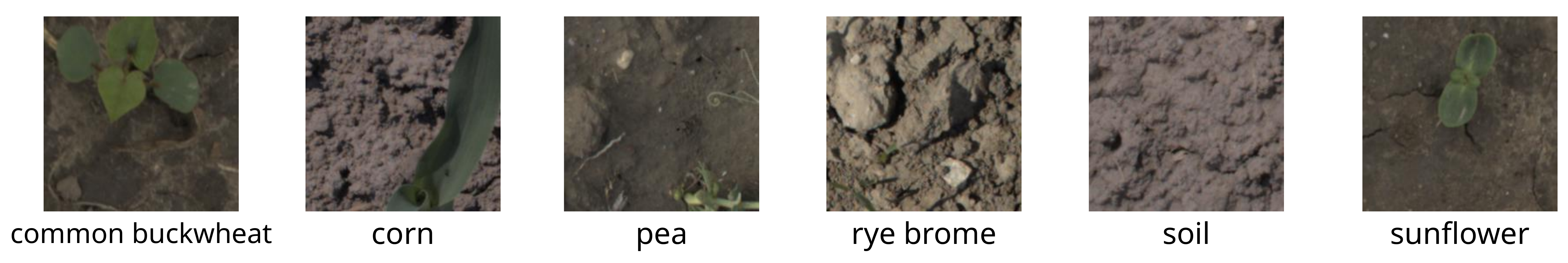}
    \caption{Some randomly chosen example images of 6 different classes. \textit{Common buckwheat} and \textit{rye brome} are weeds, while \textit{corn}, \textit{pea} and \textit{sunflower} are cultivated species.}
    \label{fig:we3ds_examples}
\end{figure*}

\subsection{Experimental Setup}

The public WE3DS dataset recently published in \citep{kitzlerWE3DSRGBDImage2023a} is originally a dataset of RGB-D images with semantic segmentation masks densely annotated into 17 plant species classes in addition to the \textit{soil} class for the background. Due to the scarcity of publicly available crop and weed classification datasets, this dataset has been transformed into a classification dataset. Discarding the depth channel, the original RGB images have been divided into non-overlapping windows of size $224 \times 224$. To each resulting image is associated a true class label which is defined as the class with the highest number of pixels in the corresponding semantically annotated mask. This results into a dataset of around 14,800 RGB images with 13 different classes, of which six random specimens are shown in \autoref{fig:we3ds_examples}. We refer the interested reader to Section 2 in the Supplementary Material for a full description of the data preparation procedure.

The database is then randomly divided into: (1) a training set ($70\%$), on which a ResNet18 classifier \citep{heDeepResidualLearning2016} is trained using default hyperparameters and pretrained weights on ImageNet \citep{dengImageNetLargeScaleHierarchical2009a}, and fixed for all experiments; the remaining $30\%$ of the data are then split into (2) a calibration set ($13.5\%$) for conformal calibration and (3) a test set ($16.5\%$) on which the conformal classifiers are evaluated. It is important to note that the choice of the base model $\mathcal{B}$ is not of great importance and is not the focal point of this study. It is for this reason, and especially to be able to study the differences among the nonconformity score functions, that we opted for a classical ResNet18 classifier which does not manifest exceptional classification performance on this task. It could have very well been replaced by a newer state-of-the-art deep classifier.

After training the ResNet18 classifier, the neural network is calibrated using each of the previously presented nonconformity score functions at the chosen confidence level of $1-\alpha = 0.9$. Then, it is used to predict sets of classes for the test images. To make sure that the obtained results are not simply due to having favorable samples of images, the calibration and test steps are repeated 1000 times, each time on a different random split of the data. The random seed of the $i^{th}$ random split, $i = 1, 2, ..., 1000$, is the same across the different nonconformity score functions so as to obtain results that are truly comparable and not simply influenced by the aleatoric uncertainty inherent to the data.

\subsection{Setting $\gamma$ and $\lambda$ for RePIP and RAPS}
For RAPS and RePIP, $k_{reg}$ is fixed at $3$ based on this specific use case's requirements. In general, we prefer not to have prediction sets with more than 3 classes: the cultivated species, a weed species and the soil. In order to choose the regularization amounts $\lambda$ and $\gamma$, we conduct a parameter sweep by testing multiple values from a manually defined grid. For each value and each method, a different conformal classifier is obtained for which we compute the efficiency and informativeness. Similarly to the experimental setup, with the aim of verifying the reliability of the estimated metrics, each conformal classifier is calibrated and tested on multiple random splits of the data.

\begin{figure}[h!]
\begin{subfigure}{0.48\textwidth}
\centering
    \includegraphics[width=\textwidth]{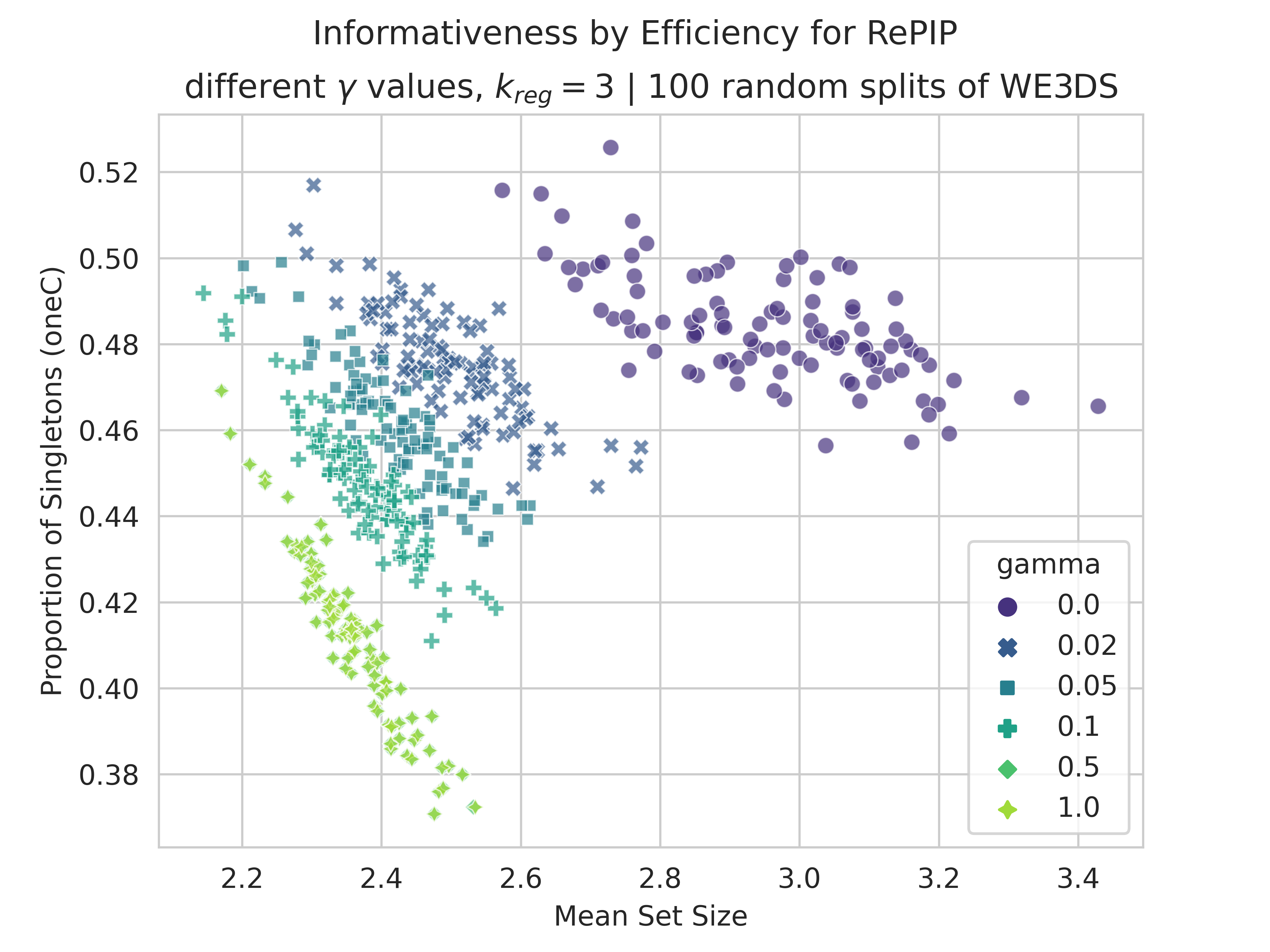}
    \caption{}
    \label{fig:gamma}
\end{subfigure}
\hfill
\begin{subfigure}{0.48\textwidth}
\centering
    \includegraphics[width=\textwidth]{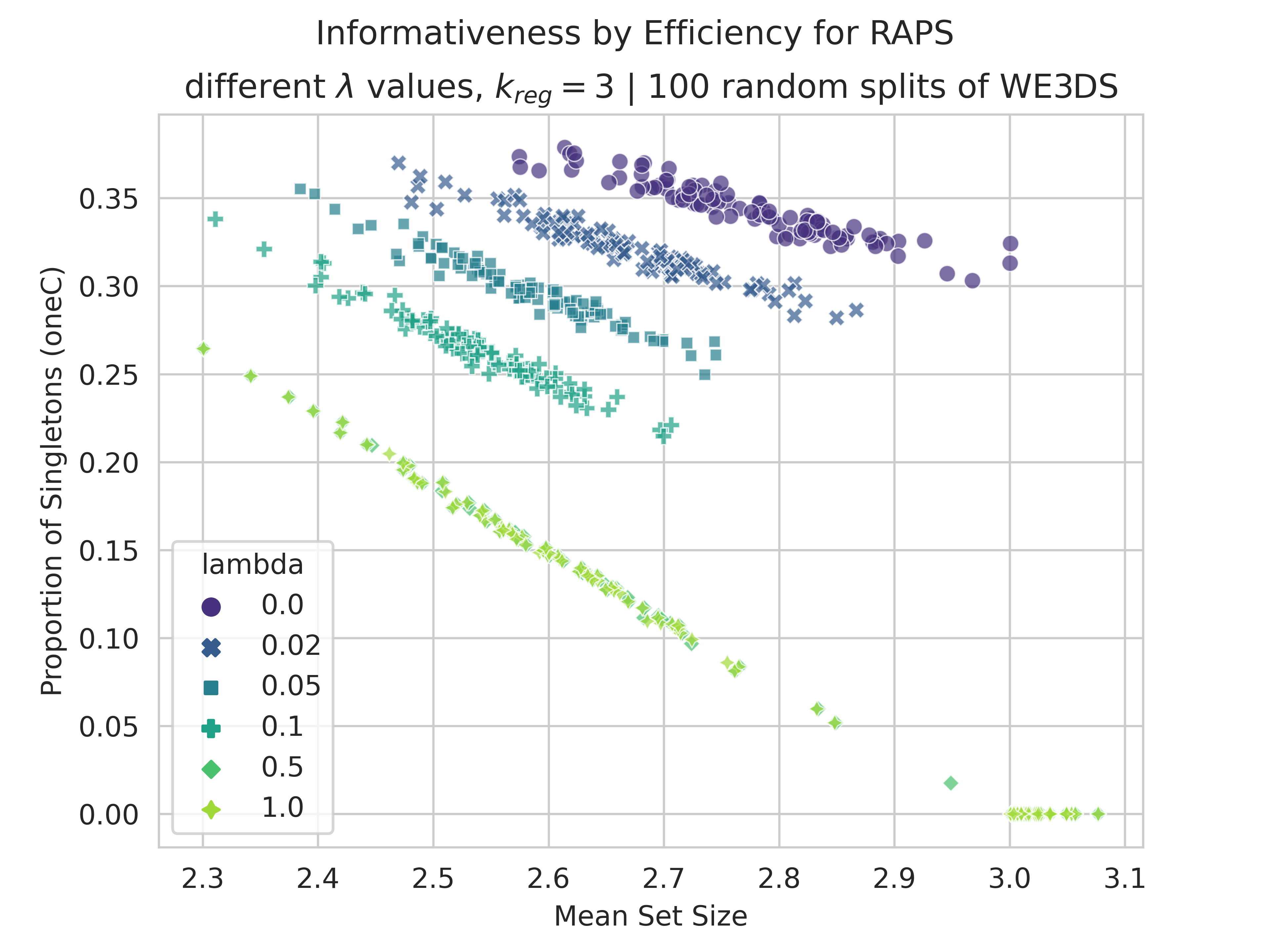}
    \caption{}
    \label{fig:lambda}
\end{subfigure}
\caption{\textit{Efficiency} and \textit{Informativeness} for different values of the regularization hyperparameters. For each value of $\gamma$ and $\lambda$, 100 different splits of the calibration and test sets are considered for more reliable results.}
\label{fig:hyperparameters}
\end{figure}

\autoref{fig:hyperparameters} shows the average set size and the proportion of singletons for each random split of the data and different values of $\gamma$ (\autoref{fig:gamma}) and $\lambda$ (\autoref{fig:lambda}). Depending on the user's preferences and the use case requirements, the optimal value can be chosen so as to place more weight on minimizing inefficiency or maximizing informativeness. 

In our case, we deem it more important to maximize the number of predicted singletons while maintaining the coverage guarantee, as it is much easier to construct decision rules when only one class is predicted. Therefore, based on the empirical results in \autoref{fig:hyperparameters}, we choose $\gamma = \lambda = 0.02$ as values for the hyperparameters. We also note that for both hyperparameters, a limit seems to be reached at $0.5$ whereby any greater value produces the same prediction sets (notice that the data points for the values $0.5$ and $1$ are overlapping).

\subsection{Results and Discussion}

\begin{figure}[htbp]
\centering
\begin{subfigure}{0.4\textwidth}
\centering
    \includegraphics[width=\textwidth]{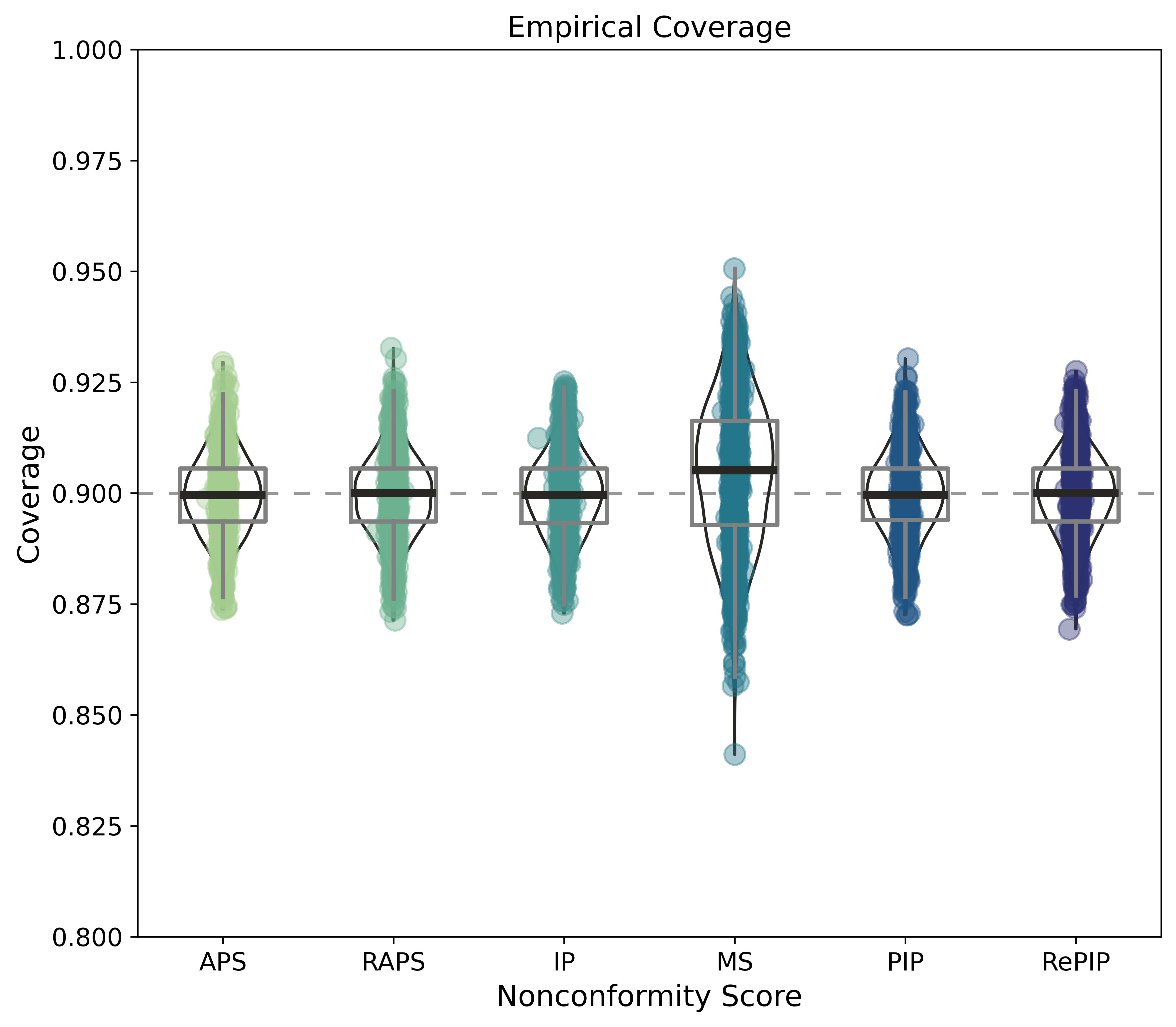}
    \caption{}
    \label{fig:coverage}
\end{subfigure}
\hfill
\begin{subfigure}{0.4\textwidth}
\centering
    \includegraphics[width=\textwidth]{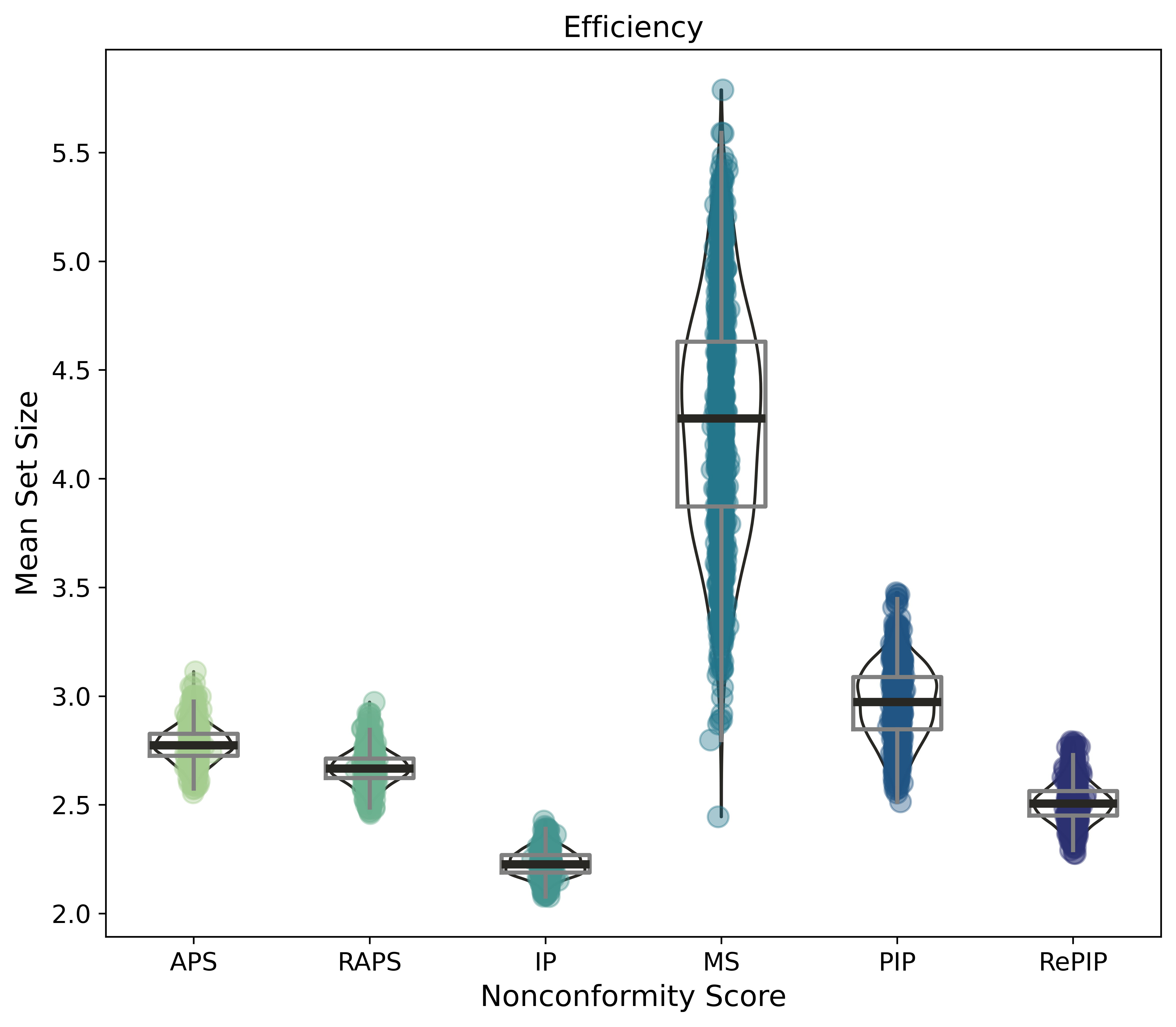}
    \caption{}
    \label{fig:efficiency}
\end{subfigure}
\hfill
\begin{subfigure}{0.4\textwidth}
\centering
    \includegraphics[width=\textwidth]{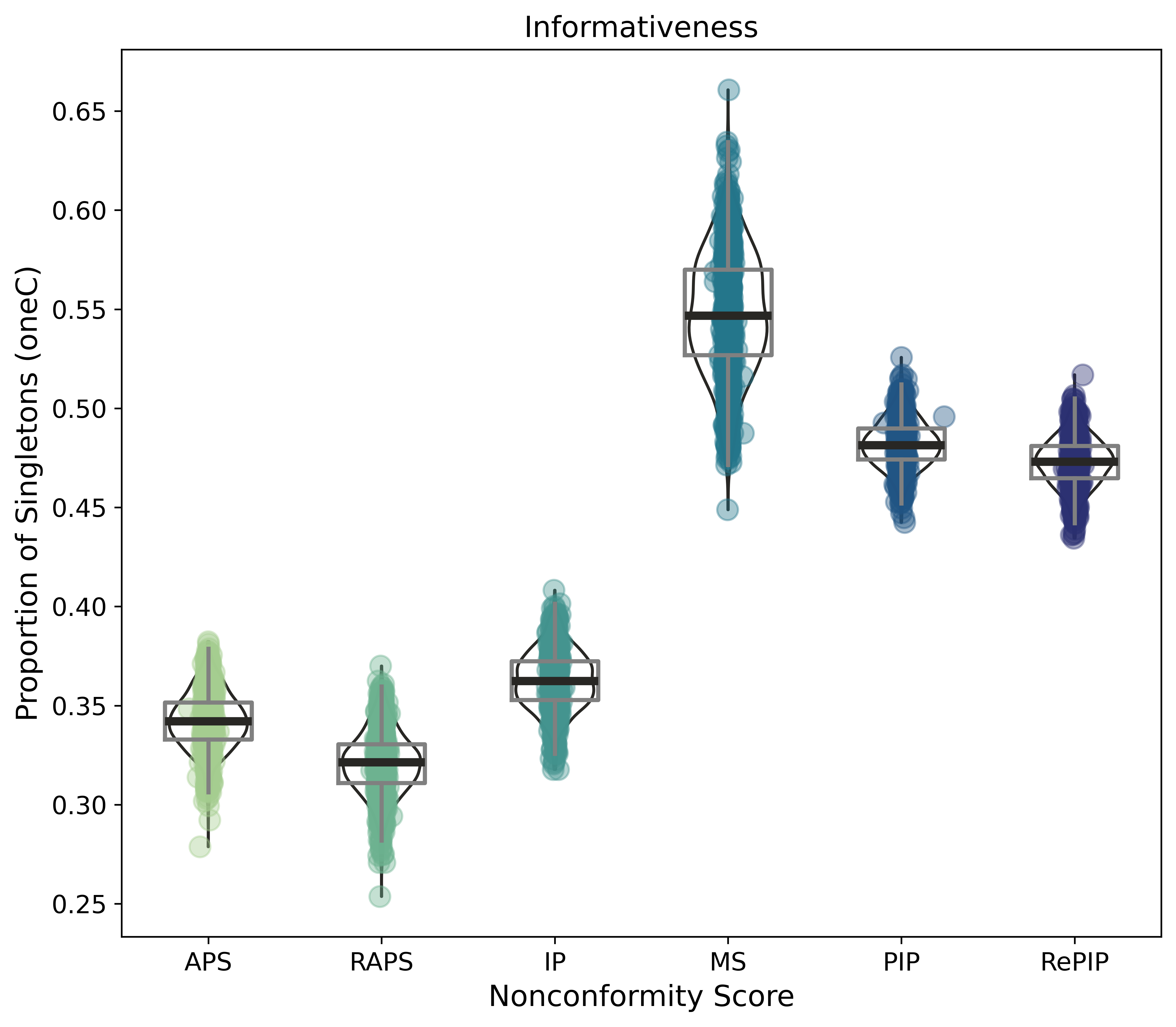}
    \caption{}
    \label{fig:informativeness}
\end{subfigure}
\caption{Violin plots of experimental results on 1000 random splits of the WE3DS classification dataset (each point is a random split): (a) \textit{Empirical Coverage} -- (b) \textit{Efficiency} (Mean Set Size) -- (c) \textit{Informativeness} (Proportion of Predicted Singletons).}
\label{fig:results}
\end{figure}
The comparison of the different models is conducted based on the efficiency and informativeness criteria. A desirable model is one that optimizes both of these criteria by producing prediction sets with small size on average and as many singletons as possible without violating the marginal coverage guarantee. \autoref{fig:results} shows the results obtained over the 1000 runs for each conformal classifier. Unsurprisingly, all the conformal classifiers are able to maintain the required $90\%$ marginal coverage guarantee on average, with MS showing a comparatively unstable behavior with respect to the other measures (\autoref{fig:coverage}).

As can be seen in \autoref{fig:efficiency}, the Hinge (IP) score leads to the smallest average set size, which is in accordance with the empirical results in \citep{johanssonModelagnosticNonconformityFunctions2017} showing that IP is the measure to use to maximize efficiency. RAPS and RePIP which are designed with efficiency in mind through the regularization term lead to slightly larger set sizes on average, with RePIP coming in second place after IP. A slight difference between APS and RAPS can be noticed. The Margin (MS) score function shows a significantly unstable behavior over the different random runs. This can be due to its deep dependence on the data it faces via the outputs of the base classifiers, an inference that can be made by comparing the divergent results in \citep{johanssonModelagnosticNonconformityFunctions2017} and \cite{aleksandrovaImpactOfModelAgnostic}. It also manifests a considerably higher average set size on average than all the other methods, a result in agreement with Johansson \textit{et al.} \citep{johanssonModelagnosticNonconformityFunctions2017}. The proposed PIP score, while exhibiting a slightly larger average set size than the other methods, is still much more efficient than the MS. This slight inefficiency is a price to pay for a considerable increase in informativeness (see~ \autoref{eq:informativeness}).


Indeed, MS manifests the highest proportion of predicted singletons, in accordance with the literature \citep{johanssonModelagnosticNonconformityFunctions2017, aleksandrovaHowNonconformityFunctions2021}, with more than $50\%$ of predicted sets being singletons, on average. This result is influenced by the estimated probabilities of the base neural network: when the base classifier assigns a much higher estimated probability to one class in comparison to the others -- that is, it is highly ``confident" in the class it predicts -- all the other classes will be considerably penalized, and thus excluded from the prediction set. This behavior is in agreement with Case 6 in \autoref{fig:softmax_examples} and \autoref{tab:scores_examples}. This behavior tends to increase the number of predicted singletons only when the base classifier $\mathcal{B}$ already has a relatively high accuracy. The other nonconformity score functions, APS, RAPS and IP, that are not explicitly concerned with informativeness, have significantly less predicted singletons. On the other hand, our proposed PIP and RePIP scores can be considered quite competitive with MS in terms of informativeness with around $50\%$ of predicted sets having size 1, and manifest better stability with regards to the data in comparison with MS. Interestingly, while the regularization via RePIP leads to considerably smaller set sizes on average, it does not decrease informativeness in any noticeable way, thus striking the required balance between the two evaluation criteria. 

In a robotic pipeline, a conformal model that satisfies the condition of guaranteed coverage under normal conditions with such a high level of singletons along with a moderate average set size (such as with PIP or RePIP) is quite attractive. While providing around half of the predictions as singletons that can readily be used to take decisions, the conformal classifier produces the remaining predictions as sets that consist of only 2 or 3 classes, on average, on which adapted decision rules can be constructed easily for autonomous agents \citep{renRobotsThatAsk2023}.


    



\section{Conclusion}
Conformal prediction is an important methodology for developing safe, deployable, machine learning systems. As long as the data faced by the model resembles, to a certain extent, the data on which it has been calibrated, the conformal model maintains the marginal coverage guarantee. Even though this marginal warranty can be strengthened, for example to provide class-conditional \cite{romanoClassificationValidAdaptive2020, angelopoulosConformalPredictionGentle2023a, vovkConditionalValidityInductive2013} or group-conditional coverage guaranties \cite{melkiGroupConditionalConformalPrediction2023a, gibbsConformalPredictionConditional2023}, it already constitutes a strong gauge of validity for machine learning models, in particular black box neural networks that do not provide such guarantees by default. The conformal envelope around any learning model can be an important step for its certification as a valid model for deployment. 

However, while any well-calibrated conformal model can provide coverage guarantees, the utility of the predictive model as a component in a larger decisional pipeline, in fully autonomous systems or human decision support systems, depends heavily on the prediction sets produced \citep{straitouriImprovingExpertPredictions2023, renRobotsThatAsk2023}. In the current work, we introduced the \textit{Penalized Inverse Probability} (PIP), and its regularized version (RePIP), with the aim of jointly optimizing the efficiency and informativeness of conformal classifiers. PIP and RePIP, mixing elements from other nonconformity score functions, provide a well-balanced hybrid behavior. The empirical results on crop and weed classification using deep neural networks show that PIP-based classifiers lead to relatively efficient prediction sets with significantly higher level of informativeness than their counterparts. Future work will continue this line of research notably by studying the behavior the different nonconformity measures on multiple datasets consisting of varying number of classes. A promising direction of exploration in safe AI is the comparison of the performance and robustness of these different nonconformity score functions under ``abnormal" conditions, for example under distribution shifts and with regards to anomalous observations.




{\small
\bibliographystyle{IEEEtranN}
\bibliography{cvpr_lib}
}


\end{document}


\maketitle

\section{Some Mathematical Properties of PIP}
The toy examples and the empirical results show that the proposed PIP nonconformity score behaves similarly to its baseline Hinge Loss (IP) measure in certain situations, while in other times its behavior resembles that of the Margin Score (MS). In fact, there is a direct relationship between these three scores, namely:

\begin{equation} \label{eq:def_pip}
    \begin{split}
        \Delta^{\text{PIP}}(y) & = \underbrace{1 - \hat{p}^y}_{\Delta^{\text{IP}}(y)} + \sum_{r = 1}^{R(y)-1} \frac{\hat{p}^{[r]}}{r} ~\mathds{1}_{\{R(y) > 1 \}} \\
    \end{split}
\end{equation}
where $R(k)$ is the rank of class $k$ after the estimated probabilities $p^1, ..., p^K$ have been sorted in decreasing order, and $\hat{p}^{[r]}$ the probability estimate of the class having rank $r$, such that $\hat{p}^k = \hat{p}^{[R(k)]}$. For any $y$ such that $R(y) > 1$, we have:
\begin{equation}
    \begin{split}
        \Delta^{\text{PIP}}(y) & = 1 - \hat{p}^y + \hat{p}^{[1]} + \sum_{r = 2}^{R(y)-1} \frac{\hat{p}^{[r]}}{r} \\ 
        & = 1 \underbrace{- \hat{p}^y + \max_{k \ne y}}_{\Delta^{\text{MS}}(y)} \hat{p}^k + \sum_{r = 2}^{R(y)-1} \frac{\hat{p}^{[r]}}{r} \\
        & = 1 + \Delta^{\text{MS}}(y) + \sum_{r = 2}^{R(y)-1} \frac{\hat{p}^{[r]}}{r} ~~~~~~ \qedsymbol \\ 
    \end{split}
\end{equation}
From these relationships, we can study the behavior of $\Delta^{\text{PIP}}(y)$ in different possible scenarios and derive some upper and lower bounds:

\begin{enumerate}
    \item Assume the case where the class of interest $y$ is the most ``certain" class. That is, $\hat{p}^y = 1$ and for all other classes $\hat{p}^k = 0, ~ k \ne y$. In such a situation, the rank $R(y)$ of $y$ will obviously be 1. In such an optimal scenario, $y$ should be given the minimal possible score. Indeed:
    \begin{equation*}
        \Delta^{\text{PIP}}(y) = \Delta^{\text{IP}}(y) = 1 - \hat{p}^y = 0
    \end{equation*}
    is a lower bound on $\Delta^{\text{PIP}}$.
    
    \item Assume the opposite scenario whereby a class $k \ne y$ is assigned the maximal probability $\hat{p}^k = 1$, which then means that $\hat{p}^y = 0$. Such a setting should be maximally penalized since the base classifier can be considered to have made a big mistake about class $y$:
    \begin{equation}
        \begin{split}
            \Delta^{\text{PIP}}(y) &= 1 + \Delta^{\text{MS}}(y) + \underbrace{\sum_{r = 2}^{R(y)-1} \frac{\hat{p}^{[r]}}{r}}_{ = 0} \\
            &= 1 + \underbrace{\max_{k \ne y} \hat{p}^k - \hat{p}^y}_{= 1} \\
            &= 2
        \end{split}
    \end{equation}
    which is an upper bound on $\Delta^{\text{PIP}}$.
    
    \item Assume the theoretical scenario where the base classifier assigns the same probability estimate to all classes. That is, $p^k = 1/K, ~ \forall k = 1,2,..., K$. In this case, the class of interest $y$ will have the same probability estimate as all other classes and its PIP score will depend only on its rank $R(y)$, which can take any value in $1, 2, ..., K$:
    \begin{equation}
        \begin{split}
            \Delta^{\text{PIP}}(y) &= 1 - \frac{1}{K} + \sum_{r = 1}^{R(y)-1} \frac{\hat{p}^{[r]}}{r} \\
            &= 1 - \frac{1}{K} + \frac{1}{K} \left(\sum_{r = 1}^{R(y)-1} \frac{1}{r} \right) \\
            &= 1 + \frac{1}{K} \left( \sum_{r = 1}^{R(y)-1} \frac{1}{r} -1 \right) \\ 
        \end{split}
    \end{equation}
    Based on this scenario, it is apparent that PIP generally guarantees assigning a different score to each class since even in the degenerate (and impossible) case when all the classes have the same probability estimates, the nonconformity score of each class will be different. 
\end{enumerate}

From these scenarios, we can further observe the ``hybrid" behavior of PIP. Indeed, from scenario (1) it is apparent that when a class $k$ has a high estimated probability $\hat{p}^k$ (close to 1), it is this probability estimate that plays the biggest role in the final score value. When this class $k$ is the class of interest $y$, then $\hat{p}^y$ will play an important role in attenuating the final score as it will be used in the IP term, i.e. the first part in \autoref{eq:def_pip}. However, when $k$ is different than the class of interest $y$ it will play a role in increasing the score assigned to $y$ as it will reside in the penalization component of the PIP score. When the base classifier is quite ``ambivalent", assigning more or less the same scores to all classes, then the most important factor impacting the final score is the rank $R(y)$ of class $y$ but which will have decreasing importance due to the inverse rank weighting in the penalization component.

\section{WE3DS Data Preparation}

The experiments in this article are conducted on the public WE3DS dataset\footnote{The original dataset can be accessed and downloaded here: \texttt{https://zenodo.org/records/5645731}} published by Kitzler et al. \citep{kitzlerWE3DSRGBDImage2023a}. Originally conceived as a dataset of densely annotated RGB-D images for semantic segmentation with 17 different plant classes and the \textit{soil} class, it has been transformed through a simple procedure into a classification dataset.
\begin{figure}[hbt!]
    \centering
    \includegraphics[width=0.5\textwidth]{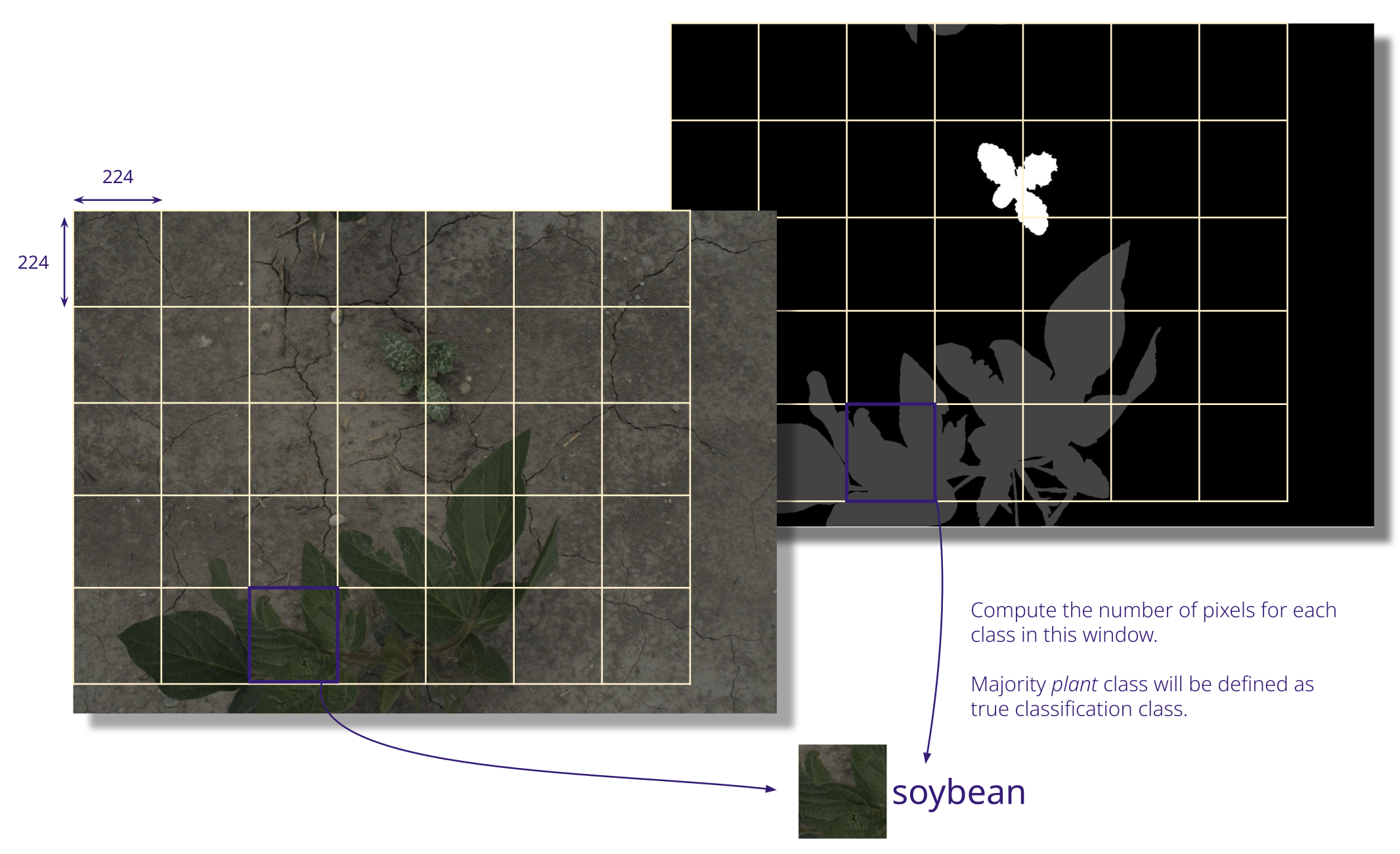}
    \caption{Demonstration of how the original WE3DS segmentation images are divided into smaller classification images.}
    \label{fig:crop_demo}
\end{figure}

As shown in \autoref{fig:crop_demo}, after discarding the depth channel, each original RGB image of size $1600 \times 1144$ is divided into non-overlapping smaller images of size $224 \times 224$. The corner regions that do not align with the cropping grid (due to the original dimensions not being perfectly divisible by $224$) are simply discarded. For each smaller image (like the one highlighted in blue in \autoref{fig:crop_demo}), its corresponding area is considered in the ground-truth semantic mask. The number of pixels in each class is counted, then a decision is taken:
\begin{enumerate}
    \item If the image contains only pixels of class \textit{soil}, then \textit{soil} is defined as the class label of the resulting classification image;
    \item If any other class exists in the image, the majority class is taken to be the true label.
\end{enumerate}

\begin{figure}
    \centering
    \includegraphics[width=0.75\linewidth]{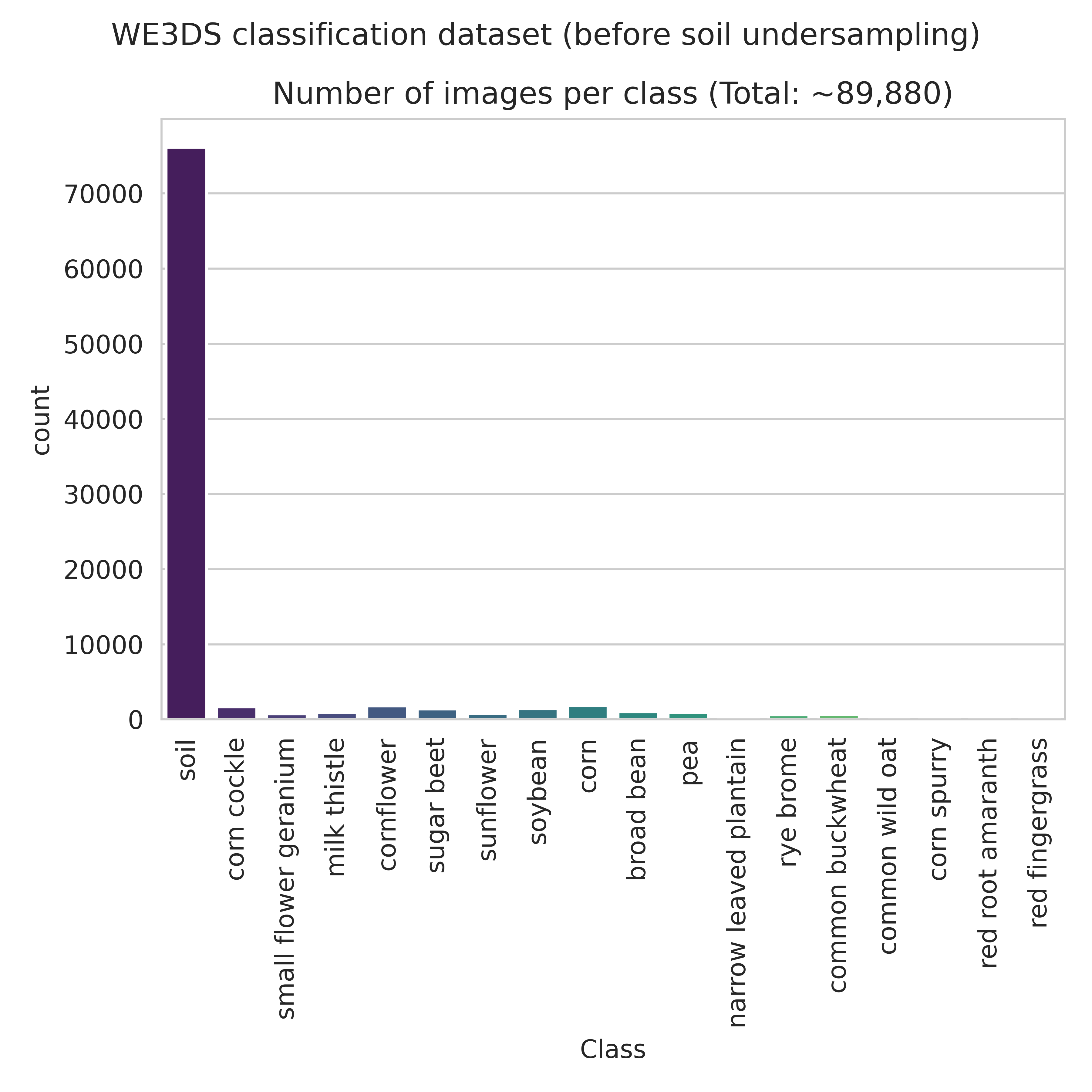}
    \caption{Count of images per class in the resulting classification dataset after window cropping from the original larger images. The \textit{soil} class overshadows all other classes.}
    \label{fig:original_distrib}
\end{figure}

\begin{figure}
    \centering
    \includegraphics[width=0.75\linewidth]{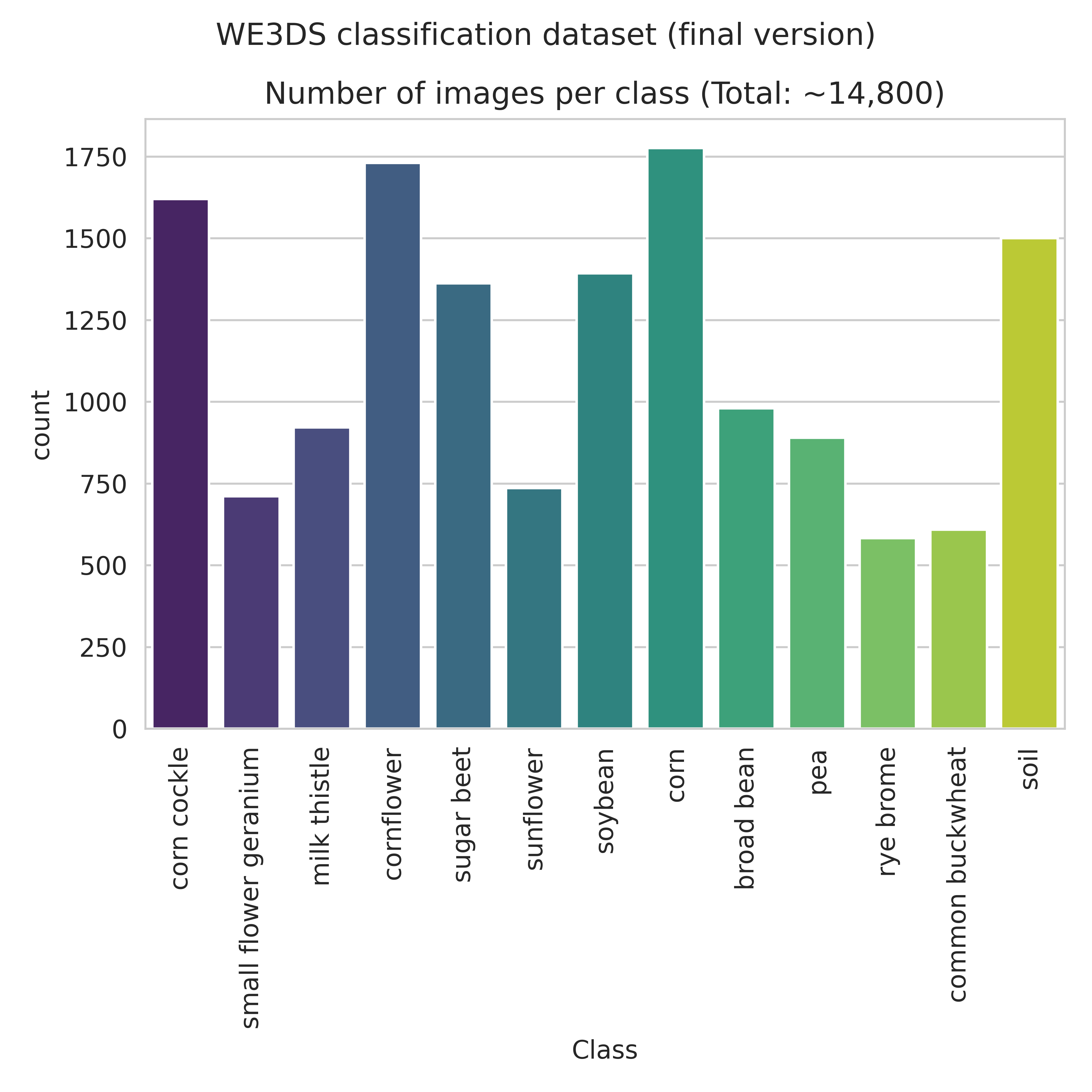}
    \caption{Count of images per class in the final classification dataset after \textit{soil} random undersampling and the dropping of the five rarest classes.}
    \label{fig:undersampled_distrib}
\end{figure}

This results in a classification dataset consisting of 89,880 images 18 different classes showing very high imbalance towards the heavily majoritarian \textit{soil} class (\autoref{fig:original_distrib}). In order to curb this imbalance problem, 1,500 images are randomly sampled from the \textit{soil} class. Additionally, the five very rare classes (\textit{corn spurry}, \textit{narrow leaved plantain}, \textit{common wild oat}, \textit{red root amaranth} and \textit{red fingergrass}) are removed, resulting in a dataset of 13 classes and 14,800 images as shown in \autoref{fig:undersampled_distrib}. This is the dataset used to conduct the experiments in the current work.

{\small
\bibliographystyle{IEEEtranN}
\bibliography{cvpr_lib}
}